\documentclass[10pt,twocolumn,twoside]{IEEEtran}
\usepackage{amsmath,graphicx}
\usepackage{amssymb}
\usepackage{graphicx}
\usepackage{graphics}
\usepackage{epstopdf}
\usepackage{amsmath}
\usepackage{algorithmicx}
\usepackage{algorithm}
\usepackage{algpseudocode}
\usepackage{multirow}
\usepackage{bm}
\usepackage{enumerate}
\usepackage{paralist}
\usepackage{array}

\begin{document}

\title{Learning Fast Sparsifying Transforms}

\author{Cristian Rusu and John Thompson\thanks{The authors are with the Institute for Digital Communications, School of Engineering, The University of Edinburgh, Scotland.
Email: \{c.rusu, john.thompson\}@ed.ac.uk. Demo source code available online at https://udrc.eng.ed.ac.uk/sites/udrc.eng.ed.ac.uk/files/attachments/demo.zip.

This work was supported by the Engineering and Physical Sciences Research Council (EPSRC) Grant number EP/K014277/1 and the MOD University Defence Research Collaboration (UDRC) in Signal Processing.}}

\maketitle

\begin{abstract}
Given a dataset, the task of learning a transform that allows sparse representations of the data bears the name of dictionary learning. In many applications, these learned dictionaries represent the data much better than the static well-known transforms (Fourier, Hadamard etc.). The main downside of learned transforms is that they lack structure and therefore they are not computationally efficient, unlike their classical counterparts. These posse several difficulties especially when using power limited hardware such as mobile devices, therefore discouraging the application of sparsity techniques in such scenarios. In this paper we construct orthogonal and non-orthogonal dictionaries that are factorized as a product of a few basic transformations. In the orthogonal case, we solve exactly the dictionary update problem for one basic transformation, which can be viewed as a generalized Givens rotation, and then propose to construct orthogonal dictionaries that are a product of these transformations, guaranteeing their fast manipulation. We also propose a method to construct fast square but non-orthogonal dictionaries that are factorized as a product of few transforms that can be viewed as a further generalization of Givens rotations to the non-orthogonal setting. We show how the proposed transforms can balance very well data representation performance and computational complexity. We also compare with classical fast and learned general and orthogonal transforms.
\end{abstract}


\section{Introduction}

Dictionary learning methods \cite{DictMagazine} represent a well-known class of algorithms that have seen many applications in signal processing \cite{SparseRep}, image processing \cite{DictImageDenosing}, wireless communications \cite{OurICC} and machine learning \cite{TaskDrivenDictionaryLearning2012}. The key idea of this approach is not to use an off-the-shelf transform like the Fourier, Hadamard or wavelet but to learn a new transform, often called an overcomplete dictionary, for a particular task (like coding and classification) from the data itself. While the dictionary learning problem is NP-hard \cite{DictLearningIsNPHard} in general, it has been extensively studied and several good algorithms to tackle it exist. Alternating minimization methods like the method of optimal directions (MOD) \cite{MOD}, K--SVD \cite{KSVD, KSVDAnalysis} and direct optimization \cite{DirectDictionary} have been shown to work well in practice and also enjoy some theoretical performance guarantees. While learning a dictionary we need to construct two objects: the dictionary and the representation of the data in the dictionary.

One problem that arises in general when using learned dictionaries is the fact that they lack any structure. This is to be compared with the previously mentioned off-the-shelf transforms that have a rich structure. This is reflected in their low computational complexity, i.e., they can be applied directly using $O(n \log n)$ computations for example \cite{FFT}. Our goal in this paper is to provide a solution to the problem of constructing fast transforms, based upon the structure of Givens rotations, learned from training data.

We first choose to study orthogonal structures since sparse reconstruction is computationally cheaper in such a dictionary: we project the data onto the column space of the dictionary and keep the largest $s$ coefficients in magnitude to obtain the provable best $s$-term approximation. Working in an $n$ dimensional feature space, this operation has complexity $O(n^2)$. In a general non-orthogonal (and even overcomplete) dictionary, special non-linear reconstruction methods such as $\ell_1$ minimization \cite{JustRelax}, greedy approaches like orthogonal matching pursuit (OMP) \cite{GreedIsGood} or variational Bayesian algorithms like approximate message passing (AMP) \cite{Donoho10112009} need to be applied. Aside from the fact that in general these methods cannot guarantee to produce best $s$-term approximations they are also computationally expensive. For example, the classical OMP has complexity $O(s n^2)$ \cite{AKSVD} and, assuming that we are looking for sparse approximations with $s \ll n$, it is in general computationally cheaper than $\ell_1$ optimization. Therefore, considering a square orthogonal dictionary is a first step in the direction of constructing a fast transform. For the analysis dictionary, recent work based on transform learning \cite{SparsifyingTransforms2013, ELL0SparsifyingTransforms2015} has been proposed. Still, notice that computing sparse representations in such a dictionary has complexity $O(n^2)$ and therefore, our goal of constructing a fast transform cannot be reached with just a general orthogonal dictionary. We make the case that our fundamental goal is to actually build a structured orthogonal dictionary such that matrix-vector multiplications with this dictionary can be achieved with less than $O(n^2)$ operations, preferably $O(n \log n)$. This connects our paper to previous work on approximating orthogonal (and symmetric) matrices \cite{FastApproximations} such that matrix-vector multiplications are computationally efficient.

When we talk about ``learning fast sparsifying transforms" we do not refer to the efficient learning procedures (although the proposed learning methods have polynomial complexity) but we refer to the transforms themselves, i.e., once we have the transform, the computational complexity of using it is low, preferably $O(n \log n)$ to perform matrix-vector multiplication.

Previous work \cite{DoubleSparsity, EfficientDictionaries, DictCirculant, ShiftInvariantDictLearning, Rusu2013, DoublySparsifyingTransform2013, SulamOZE16} in the literature has already proposed various structured dictionaries to cope with the high computational complexity of learned transforms. Previous work also dealt with the construction of structured orthogonal dictionaries. Specifically, \cite{DictHouseholder} proposed to build an orthogonal dictionary composed of a product of a few Householder reflectors. In this fashion, the computational complexity of the dictionary is controlled and a trade-off between representation performance and computational complexity is shown.

Learned dictionaries with low computational complexity can bridge the gap between the classical transforms that are preferred especially in power limited hardware (or battery operated devices) and the overcomplete, computationally cumbersome, learned dictionaries that provide state-of-the-art performance in many machine learning tasks. The contribution of this paper is two fold.

First, we consider the problem of constructing an orthogonal dictionary as a product of a given number of generalized Givens rotations. We start by showing the optimum solution to the dictionary learning problem when the dictionary is a single generalized Givens rotation and then move to expand on this result and propose an algorithm that sequentially builds a product of generalized Givens rotations to act as a dictionary for sparse representations. Each step of the algorithm solves exactly the proposed optimization problem and therefore we can guarantee that it monotonically converges to a local minimum. We show numerically that the fast dictionaries proposed in this paper outperform those based on Householder reflectors \cite{DictHouseholder} in terms of representation error, for the same computational complexity. 

Second, based on a structure similar to the generalized Givens rotation we then propose a learning method that constructs square, non-orthogonal, computationally efficient dictionaries. In order to construct the dictionary we again solve exactly a series of optimization problems. Unfortunately we cannot prove the monotonic convergence of the algorithm since the sparse reconstruction step, based in this paper on OMP, cannot guarantee in general a monotonic reduction in our objective function. Still, we are able to show that these fast non-orthogonal transforms perform very well, better than their orthogonal counterparts.

In the results section we compare the proposed methods among each other and to previously proposed dictionary learning methods in the literature. We show that the methods proposed in this paper provide a clear trade-off between representation performance and computational complexity. Interestingly, we are able to provide numerical examples where the proposed fast orthogonal dictionaries have higher computational efficiency and provide better representation performance than the well-known discrete cosine transform (DCT), the transform at the heart of the jpeg compression standard \cite{jpeg1992}.


\section{A brief description of dictionary learning optimization problems}

Given a real dataset $\mathbf{Y} \in \mathbb{R}^{n \times N}$ and sparsity level $s$, the general dictionary learning problem is to produce the factorization $\mathbf{Y} \approx \mathbf{DX}$ given by the optimization problem:
\begin{equation}
\begin{aligned}
& \underset{\mathbf{D},\ \mathbf{X}}{\text{minimize}} & & \|\mathbf{Y}-\mathbf{DX}\|_F^2 \\
& \text{subject to} &  & \|\mathbf{x}_i\|_{0} \leq s,\ 1 \leq i \leq N \\
& & & \| \mathbf{d}_j \|_2 = 1, \ 1 \leq j \leq n,
\end{aligned}
\label{eq:dictionaryGeneral}
\end{equation}
where the objective function describes the Frobenius norm representation error achieved by the square dictionary $\mathbf{D} \in \mathbb{R}^{n \times n}$ with the sparse representations $\mathbf{X} \in \mathbb{R}^{n \times N}$ whose columns are subject to the $\ell_0$ pseudo-norm $\| \mathbf{x}_i \|_0$ (the number of non-zero elements of columns $\mathbf{x}_i$). To avoid trivial solutions, the dimensions obey $s \ll n \ll N$. Several algorithms that work very well in practice exist \cite{MOD} \cite{KSVD} \cite{AKSVD} to solve this factorization problem. Their approach, and the one we also adopt in this paper, is to keep the dictionary fixed and update the representations and then reverse the roles by updating the dictionary with the representations fixed. This alternating minimization approach proves to work very well experimentally \cite{MOD, KSVD} and allows some theoretical insights \cite{AlternatingMinimization2014}.

In this paper we also consider the dictionary learning problem \eqref{eq:dictionaryGeneral} with an orthogonal dictionary $\mathbf{Q} \in \mathbb{R}^{n \times n}$ \cite{ICIPConf} \cite{SOT} \cite{EMAlgorithm} \cite{2016Schutze}. The orthogonal dictionary learning problem (which we call in this paper Q--DLA) \cite{OrthoDictionary} is formulated as:
\begin{equation}
\begin{aligned}
& \underset{\mathbf{Q},\ \mathbf{X}; \ \mathbf{QQ}^T = \mathbf{Q}^T\mathbf{Q} = \mathbf{I}}{\text{minimize}} & & \|\mathbf{Y}-\mathbf{QX}\|_F^2 \\
& \text{\ \ \ \ \ \ subject to} &  & \|\mathbf{x}_i\|_{0} \leq s,\ 1 \leq i \leq N.
\end{aligned}
\label{eq:dictionaryOrtho}
\end{equation}

Since the dictionary $\mathbf{Q}$ is orthogonal, the construction of $\mathbf{X}$ no longer involves $\ell_1$ \cite{JustRelax}, OMP \cite{GreedIsGood} or AMP \cite{Donoho10112009} approaches as in \eqref{eq:dictionaryGeneral}, but reduces to $\mathbf{X} = \mathcal{T}_s(\mathbf{Q}^T\mathbf{Y})$, where $\mathcal{T}_s()$ is an operator that given an input vector zeros all entries except the largest $s$ in magnitude and given an input matrix applies the same operation on each column in turn. To solve \eqref{eq:dictionaryOrtho} for variable $\mathbf{Q}$ and fixed $\mathbf{X}$, a problem also known as the orthogonal Procrustes problem \cite{Proc}, a closed form solution $\mathbf{Q} = \mathbf{UV}^T$ is given by the singular value decomposition of $\mathbf{YX}^T = \mathbf{U\Sigma V}^T$.

\section{A building block for fast transforms}

For indices $(i,j),\ j > i$ and variables $p,q,r,t \in \mathbb{R}$ let us define the basic transform, which we call an R-transform:
\begin{equation}
\mathbf{R}_{ij} = \begin{bmatrix} 
\mathbf{I}_{i-1} &  &  &  & \\
& p &  & r & \\
& & \mathbf{I}_{j-i-1} & & \\
& q & & t & \\
& & & & \mathbf{I}_{n-j} \\
\end{bmatrix} \in \mathbb{R}^{n \times n},
\label{eq:Rtransform}
\end{equation}
where we have denoted $\mathbf{I}_i$ as the identity matrix of size $i$. For simplicity, we denote the non-zero part of $\mathbf{R}_{ij}$ as
 \begin{equation}
 \mathbf{\tilde{R}}_{ij}  = \begin{bmatrix}
 p & r \\
 q & t
 \end{bmatrix} \in \mathbb{R}^{2 \times 2}.
 \label{eq:localstructureR}
 \end{equation}
  A right side multiplication between a R-transform and a matrix $\mathbf{X} \in \mathbb{R}^{n \times N}$ operates only rows $i$ and $j$ as
  \begin{equation}
  \begin{split}
  \mathbf{R}_{ij} \mathbf{X} & = [\begin{matrix}  \mathbf{x}_1 & \dots & 
  p \mathbf{x}_i + r \mathbf{x}_j &
  \dots & \end{matrix} \\
  & \qquad\qquad\qquad \begin{matrix} \dots & q \mathbf{x}_i + t \mathbf{x}_j &
  \dots & \mathbf{x}_n]
  \end{matrix}^T,
  \end{split}
  \label{eq:multiply}
  \end{equation}
  where $\mathbf{x}_i^T$ is the $i^\text{th}$ row of $\mathbf{X}$. The number of operations needed for this task is only $6N$. Left and right multiplications with a R-transform (or its transpose) are therefore computationally efficient. We use this matrix structure as a basic building block for the transforms learned in this paper.
 
\noindent \textbf{Remark 1.} Every matrix $\mathbf{D} \in \mathbb{R}^{n \times n}$ can be written as a product of at most $\lceil n^2 - \frac{n}{2} \rceil$ R-transforms. Therefore, we can consider the R-transforms as fundamental building blocks for all square transforms $\mathbf{D}$.

\noindent \textit{Proof.} Consider the singular value decomposition $\mathbf{D} = \mathbf{U \Sigma V}^T$. Each $\mathbf{U}$ and $\mathbf{V}$ can be factored as a product of ${n \choose 2}$ Givens rotations \cite{Golub1996} which are all in fact constrained R-transforms (with $p = t = c$ and $r = -q = d$ for some given $c$ and $d$ such that $c^2 + d^2 = 1$) and a diagonal matrix containing only $\{\pm 1\}$ entries. While the diagonal $\mathbf{\Sigma}$ can be factored as a product of $\lceil \frac{n}{2} \rceil$ diagonal R-transforms. $\hfill \blacksquare$

In this paper we will be interested to use $\mathbf{R}_{ij}$ in least squares problems with the objective function as:
\begin{equation}
	\begin{aligned}
	& \| \mathbf{Y} -  \mathbf{R}_{ij} \mathbf{X} \|_F^2 = \| \mathbf{Y} \|_F^2 +  \| \mathbf{X} \|_F^2 - 2\text{tr}(\mathbf{Z}) -  \\ 
	& \left\| \begin{bmatrix} \mathbf{y}_i^T \\ \mathbf{y}_j^T \end{bmatrix} \right\|_F^2 \! - \! \left\| \begin{bmatrix} \mathbf{x}_i^T \\ \mathbf{x}_j^T \end{bmatrix} \right\|_F^2 \! + \! 2\text{tr}(\mathbf{Z}_{ \{i,j\} }) \! + \! \left\| \begin{bmatrix} \mathbf{y}_i^T \\ \mathbf{y}_j^T \end{bmatrix} \! - \!
	\mathbf{\tilde{R}}_{ij}
	\begin{bmatrix} \mathbf{x}_i^T \\ \mathbf{x}_j^T \end{bmatrix} \right\|_F^2.
	\end{aligned}
	\label{eq:theRexpanded}
\end{equation}
For simplicity of exposition we have defined
\begin{equation}
\mathbf{Z} = \mathbf{YX}^T,\mathbf{Z}_{\{i,j\}} = \begin{bmatrix} Z_{ii} & Z_{ij}  \\ Z_{ji} & Z_{jj} \end{bmatrix} \in \mathbb{R}^{2 \times 2},Z_{ij} = \mathbf{y}_i^T \mathbf{x}_j,
\end{equation}
where $\mathbf{y}_i^T$ and $\mathbf{x}_i^T$ are the $i^\text{th}$ rows of $\mathbf{Y}$ and $\mathbf{X}$, respectively.

We now introduce learning methods to create computationally efficient orthogonal and non-orthogonal dictionaries.

\section{A method for designing fast orthogonal transforms: G$_m$--DLA}

In this section we propose a method called G$_m$--DLA to learn orthogonal dictionaries that are factorized as a product of $m$ G-transforms (constrained R-transforms).

\subsection{An overview of G-transforms}

We call $\mathbf{G}_{ij}$ a G-transform, an orthogonal constrained R-transform \eqref{eq:Rtransform} parameterized only by $c, d \in \mathbb{R}$ with $c^2 + d^2 = 1$, and the indices $(i,j),\ i \neq j$ such that the non-zero part of $\mathbf{G}_{ij}$, corresponding to \eqref{eq:localstructureR}, is given by
\begin{equation}
	\mathbf{\tilde{G}}_{ij}  \in \left\{ \begin{bmatrix}
					c & d \\
					-d & c
	\end{bmatrix},\ \begin{bmatrix}
	c & d \\
	d & -c
	\end{bmatrix} \right\}.
	\label{eq:localstructure}
\end{equation}
Classically, a Givens rotation is a matrix as in \eqref{eq:Rtransform} with $\mathbf{\tilde{G}}_{ij}  = \begin{bmatrix}
c & d \\
-d & c
\end{bmatrix}$ such that $\det(\mathbf{G}_{ij}) = 1$, i.e., proper rotation matrices are orthogonal matrices with determinant one. These rotations are important since any orthogonal dictionary of size $n \times n$ can be factorized in a product of ${n \choose 2}$ Givens rotations \cite{Golub1996}. In this paper, since we are interested in the computational complexity of these structures, we allow both options in \eqref{eq:localstructure} that fully characterize all $2 \times 2$ real orthogonal matrices -- these structures are discussed in \cite[Chapter~2.1]{MatrixAnalysis}. With $\mathbf{\tilde{G}}_{ij}  = \begin{bmatrix}
c & d \\
d & -c
\end{bmatrix}$ the G-transform in \eqref{eq:Rtransform} is in fact a Householder reflector $\mathbf{G}_{ij} = \mathbf{I} - 2\mathbf{g}_{ij}\mathbf{g}_{ij}^T$ where $\mathbf{g}_{ij} \in \mathbb{R}^n, \ \| \mathbf{g}_{ij} \|_2 = 1,$ has all entries equal to zero except for the $i^\text{th}$ and $j^\text{th}$ entries that are $\sqrt{0.5(1-c)}$ and $-\text{sign}(d)\sqrt{0.5(1+c)}$, respectively -- one might call this a ``Givens reflector'' to highlight its distinguishing sparse structure. Givens rotations have been previously used in matrix factorization applications \cite{SparseMatrixTransform2011, MultiresolutionMatrixFactorization2014}.

\subsection{One G-transform as a dictionary}

Consider now the dictionary learning problem in \eqref{eq:dictionaryOrtho}. Let us keep the sparse representations $\mathbf{X}$ fixed and consider a single G-transform as a dictionary. We reach the following
		\begin{equation}
			\begin{aligned}
\underset{(i,j),\ \mathbf{\tilde{G}}_{ij}}{\text{minimize}} \ \  \| \mathbf{Y} - \mathbf{G}_{ij} \mathbf{X} \|_F^2.
		\label{eq:GivensLearning}
			\end{aligned}
		\end{equation}

When indices $(i,j)$ are fixed, the problem reduces to constructing $\mathbf{\tilde{G}}_{ij}$, a constrained two dimensional optimization problem. To select the indices $(i,j)$, among the ${n \choose 2}$ possibilities, an appropriate strategy needs to be defined. We detail next how to deal with these two problems to provide an overall solution for \eqref{eq:GivensLearning}.

\noindent \textbf{To solve \eqref{eq:GivensLearning} for the fixed coordinates $(i,j)$} we reach the optimization problem
\begin{equation}
\underset{\mathbf{\tilde{G}}_{ij};\ \mathbf{\tilde{G}}_{ij}^T\mathbf{\tilde{G}}_{ij}= \mathbf{\tilde{G}}_{ij}\mathbf{\tilde{G}}_{ij}^T =  \mathbf{I}}{\text{minimize}} 
\left\| \begin{bmatrix} \mathbf{y}_i^T \\ \mathbf{y}_j^T \end{bmatrix} -
\mathbf{\tilde{G}}_{ij}
\begin{bmatrix} \mathbf{x}_i^T \\ \mathbf{x}_j^T \end{bmatrix} \right\|_F^2.
\label{eq:GivensLearning2}
\end{equation}
This is a two dimensional Procrustes problem \cite{Proc} whose optimum solution is $\mathbf{\tilde{G}}_{ij} = \mathbf{UV}^T$ where $ \mathbf{Z}_{\{i,j\}} = \mathbf{U\Sigma V}^T$. It has been shown in \cite{DictHouseholder} that the reduction in the objective function of \eqref{eq:GivensLearning2} when considering an orthogonal dictionary $\mathbf{G}_{ij}$ given by the Procrustes solution is
\begin{equation}
\begin{aligned}
& \left\| \begin{bmatrix} \mathbf{y}_i^T \\ \mathbf{y}_j^T \end{bmatrix} -
\mathbf{\tilde{G}}_{ij}
\begin{bmatrix} \mathbf{x}_i^T \\ \mathbf{x}_j^T \end{bmatrix} \right\|_F^2 \\ 
& \quad = \left\| \begin{bmatrix} \mathbf{y}_i^T \\ \mathbf{y}_j^T \end{bmatrix} \right\|_F^2 + \left\| \begin{bmatrix} \mathbf{x}_i^T \\ \mathbf{x}_j^T \end{bmatrix} \right\|_F^2 \! \! - \! \! 2\text{tr} \left(\! \! \mathbf{G}_{ij}^T \begin{bmatrix} \mathbf{y}_i^T \\ \mathbf{y}_j^T \end{bmatrix} \begin{bmatrix} \mathbf{x}_i^T \\ \mathbf{x}_j^T \end{bmatrix}^T \right) \\
&  \quad = \left\| \begin{bmatrix} \mathbf{y}_i^T \\ \mathbf{y}_j^T \end{bmatrix} \right\|_F^2 + \left\| \begin{bmatrix} \mathbf{x}_i^T \\ \mathbf{x}_j^T \end{bmatrix} \right\|_F^2 - 2 \| \mathbf{Z}_{ \{i,j\} } \|_*,
\end{aligned}
\label{eq:objectivewithQ}
\end{equation}
where $\| \mathbf{Z}_{ \{i,j\} } \|_*$ is the nuclear norm of $\mathbf{Z}_{ \{i,j\} }$, i.e., the sum of its singular values.

\noindent \textbf{Choosing $(i,j)$ in \eqref{eq:GivensLearning}} requires a closer look at its objective function \eqref{eq:theRexpanded} for $\mathbf{\tilde{R}}_{ij}  = \mathbf{\tilde{G}}_{ij}$, the constrained G-transform structure. Using \eqref{eq:objectivewithQ} we can state a result in the special case of a G-transform. We need both because for any indices $(i,j)$ the reduction in the objective function invokes the nuclear norm, while for the other indices the reduction invokes the trace. We can analyze the two objective function values separately because the Frobenius norm is elementwise and as such also blockwise. Therefore, the objective of \eqref{eq:GivensLearning} is
\begin{equation}
	\begin{aligned}
	& \| \mathbf{Y} - \mathbf{G}_{ij}\mathbf{X} \|_F^2 = \| \mathbf{Y} \|_F^2 + \| \mathbf{X} \|_F^2 - 2\text{tr}(\mathbf{Z}) - 2C_{ij}, \\
	& \text{where } C_{ij} = \| \mathbf{Z}_{\{ i,j\}} \|_* - \text{tr}(\mathbf{Z}_{\{ i,j\}}).
	\end{aligned}
	\label{eq:detailedobjective}
\end{equation}
Since we want to minimize this quantity, the choice of indices needs to be made as follows
\begin{equation}
	(i^\star, j^\star) = \underset{(i, j),\ j > i}{\arg \max}\ C_{ij},
	\label{eq:score}
\end{equation}
and then solve a Procrustes problem \cite{Proc} to construct $\mathbf{\tilde{G}}_{i^\star j^\star}$.

These $(i^\star, j^\star)$ values are the optimum indices that lead to the maximal reduction in the objective function of \eqref{eq:GivensLearning}. The expression in \eqref{eq:score} is computationally cheap given that $\mathbf{Z}_{\{ i,j\}}$ is a $2 \times 2$ real matrix. Its trace is trivial to compute $\text{tr}( \mathbf{Z}_{\{i,j\}} ) = Z_{ii} + Z_{jj}$ (one addition operation) while the singular values of $\mathbf{Z}_{\{i,j\}}$ can be explicitly computed as
\begin{equation}
	\sigma_{1,2} \! \! = \! \! \sqrt{\frac{1}{2} \left(\!  \| \mathbf{Z}_{\{ i,j\}} \|_F^2 \! \pm \! \sqrt{\|\mathbf{Z}_{\{ i,j\}}\|_F^4 \! - \! 4 \det(\mathbf{Z}_{\{ i,j\}})^2} \! \right) }.
	\label{eq:singularvalues}
\end{equation}
Therefore, the full singular value decomposition can be avoided and the sum of the singular values from \eqref{eq:singularvalues} can be computed in only 23 operations (three of which are taking square roots). The cost of computing $C_{ij}$ for all indices $(i,j),\ j > i,$ is $25 \frac{n(n-1)}{2}$ operations. The computational burden is still dominated by constructing $\mathbf{Z} = \mathbf{YX}^T$ which takes $2snN$ operations.

\noindent \textbf{Remark 2.} Notice that $C_{ij} \geq 0$ always. In general, this is because the sum of the singular values of any matrix $\mathbf{Z}$ of size $n \times n$ is always greater than the sum of its eigenvalues. To see this, use the singular value decomposition of $\mathbf{Z} = \mathbf{U \Sigma V}^T, \mathbf{\Sigma} = \text{diag}(\mathbf{\sigma})$, and develop:
\begin{equation}
	\text{tr}(\mathbf{Z}) = \text{tr}(\mathbf{\Sigma V}^T \mathbf{ U})
			  = \sum_{k=1}^n \sigma_{k} \Delta_{kk} \leq \sum_{k=1}^n \sigma_k  = \| \mathbf{Z} \|_*,
	\label{eq:traceandnuclear}
\end{equation}
where we have use the circular property of the trace and $\mathbf{\Delta} = \mathbf{V}^T \mathbf{U}$ where $\Delta_{kk}$ are its diagonal entries which obey $| \Delta_{kk} | \leq 1$ since both $\mathbf{U}$ and $\mathbf{V}$ are orthogonal and their entries are sub-unitary in magnitude. Therefore, in our particular case, we have that $C_{ij} = 0$ when $\mathbf{Z}_{\{i,j\}}$ is symmetric and positive semidefinite (we have that $\mathbf{\Delta} = \mathbf{I}$ in \eqref{eq:traceandnuclear} and therefore $\text{tr}(\mathbf{Z}_{\{i,j\}}) = \| \mathbf{Z}_{\{i,j\}} \|_*$). If we have that $C_{ij} = 0$ for all $i$ and $j$ then no G-transform can reduce the objective function in \eqref{eq:GivensLearning} and therefore the solution is $\mathbf{G}_{ij} = \mathbf{I}$. $\hfill \blacksquare$

\noindent \textbf{Remark 3.} We can extend the G-transform to multiple indices. For example, if we consider three coordinates then $\mathbf{G}_{ijk} \in \mathbb{R}^{n \times n}$ has the non-zero orthogonal block $\mathbf{\tilde{G}}_{ijk} \in \mathbb{R}^{3 \times 3}$. For a transform over $q$ indices there are ${n \choose q}$ such blocks and its matrix-vector multiplication takes $(2q-1)q$ operations.$\hfill \blacksquare$

\noindent \textbf{Remark 4.} There are some connections between the Householder \cite{DictHouseholder} and the G-transform approaches. As previously explained, when $\mathbf{\tilde{G}}_{ij}  = \begin{bmatrix}
c & d \\
d & -c
\end{bmatrix}$ the G-transform can also be viewed as a Householder reflector, i.e., $\mathbf{G}_{ij} = \mathbf{I} - 2\mathbf{g}_{ij}\mathbf{g}_{ij}^T$ where $\mathbf{g}_{ij} \in \mathbb{R}^n$ is a 2-sparse vector. Following results from \cite{DictHouseholder} we can also write
	\begin{equation}
		\begin{aligned}
	\| \mathbf{Y} - \mathbf{G}_{ij}\mathbf{X} \|_F^2 = & \| \mathbf{Y} \|_F^2 + \|  \mathbf{X} \|_F^2 - 2\text{tr}(\mathbf{YX}^T)  \\
	& \ \ \ \ \ \ \ \ + 2\mathbf{g}_{ij}^T (\mathbf{YX}^T + \mathbf{XY}^T) \mathbf{g}_{ij},
		\end{aligned}
	\end{equation}
	which, together with \eqref{eq:detailedobjective}, leads to $\mathbf{g}_{ij}^T (\mathbf{YX}^T + \mathbf{XY}^T) \mathbf{g}_{ij} = -C_{ij}$. This means that choosing to maximize $C_{ij}$ in \eqref{eq:detailedobjective} is equivalent to computing an eigenvector of $\mathbf{YX}^T + \mathbf{XY}^T$ of sparsity two associated with a negative eigenvalue.
	
	There are also some differences between the two approaches. For example, matrix-vector multiplication with a G-transform $\mathbf{G}_{ij} \mathbf{x}$ takes 6 operations but when using the Householder structure $\mathbf{G}_{ij} \mathbf{x} = (\mathbf{I} - 2\mathbf{g}_{ij}\mathbf{g}_{ij}^T)\mathbf{x} = \mathbf{x} - 2(\mathbf{g}_{ij}^T \mathbf{x}) \mathbf{g}_{ij}$ takes 8 operations (4 operations to compute the constant $C = 2\mathbf{g}_{ij}^T \mathbf{x}$, 2 operations to compute the 2-sparse vector $\mathbf{z} = C\mathbf{g}_{ij}$ and 2 operations to compute the final result $\mathbf{x} - \mathbf{z}$). Therefore, the G-transform structure is computationally preferable to the Householder structure. Each Householder reflector has $n-1$ (because of the orthogonality constraint) degrees of freedom while each G-transform has only $1$ (the angle $\theta \in [0, 2\pi]$ for which $c = \cos \theta$ and $d = \sin \theta$) plus 1 bit (the choice of the rotation or reflector in \eqref{eq:localstructure}). $\hfill \blacksquare$

This concludes our discussion for the single G-transform case. Notice that the solution outlined in this section solves \eqref{eq:GivensLearning} exactly, i.e., it finds the optimum G-transform.

\subsection{A method for designing fast orthogonal transforms: G$_m$--DLA}

In this paper we propose to construct an orthogonal transform $\mathbf{U} \in \mathbb{R}^{n \times n}$ with the following structure:
\begin{equation}
\mathbf{U} = \mathbf{G}_{i_m j_m} \dots \mathbf{G}_{i_2 j_2} \mathbf{G}_{i_1 j_1}.
\label{eq:factorization}
\end{equation}
The value of $m$ is a choice of the user. For example, if we choose $m$ to be $O(n \log n)$ the transform $\mathbf{U}$ can be computed in $O(n \log n)$ computational complexity -- similar to the classical fast transforms. The goal of this section is to propose a learning method that constructs such a transform.

We fix the representations $\mathbf{X}$ and all G-transforms in \eqref{eq:factorization} except for the $k^\text{th}$, denoted as $\mathbf{G}_{i_k j_k}$. To optimize the dictionary $\mathbf{U}$ only for this transform we reach the objective function
\begin{equation}
\begin{aligned}
\| \mathbf{Y} & -  \mathbf{UX} \|_F^2 = \| \mathbf{Y} - \mathbf{G}_{i_m j_m} \dots \mathbf{G}_{i_1 j_1} \mathbf{X} \|_F^2 \\
= & \| \mathbf{G}_{i_{k+1} j_{k+1}}^T \dots \mathbf{G}_{i_m j_m}^T \mathbf{Y}  - \mathbf{G}_{i_k j_k} \dots \mathbf{G}_{i_1 j_1} \mathbf{X} \|_F^2 \\
= & \| \mathbf{Y}_k  - \mathbf{G}_{i_k j_k} \mathbf{X}_k \|_F^2,
\end{aligned}
\label{eq:solvemultipleGivens}
\end{equation}
where we have used the fact that multiplication by any orthogonal transform preserves the Frobenius norm. For simplicity we have denoted $\mathbf{Y}_k$ and $\mathbf{X}_k$ the known quantities in \eqref{eq:solvemultipleGivens} and therefore $\mathbf{Z}_k = \mathbf{Y}_k \mathbf{X}_k^T$.

Notice that we have reduced the problem to the formulation in \eqref{eq:GivensLearning} whose full solution is outlined in the previous section. We can apply this procedure for all G-transforms in the product of $\mathbf{U}$ and therefore a full update procedure presents itself: we will sequentially update each transform and then the sparse representations until convergence. The full procedure we propose, called G$_m$--DLA, is detailed in Algorithm 1.

\noindent\textbf{The initialization of G$_m$--DLA} uses a known construction. It has been shown experimentally in the past \cite{DictInit}, that a good initial orthogonal dictionary is to choose $\mathbf{U}$ from the singular value decomposition of the dataset $\mathbf{Y} = \mathbf{U\Sigma V}^T$. We can also provide a theoretical argument for this choice. Consider that
\begin{equation}
\mathbf{X} = \mathcal{T}_s(\mathbf{U}^T \mathbf{Y}) = \mathcal{T}_s(\mathbf{U}^T \mathbf{U \Sigma V}^T) = \mathcal{T}_s(\mathbf{\Sigma V}^T).
\end{equation}
A sub-optimal choice is to assume that the operator $\mathcal{T}_s$ keeps only the first $s$ rows of $\mathbf{\Sigma V}^T$, i.e., $\mathbf{X} = \mathbf{\Sigma}_s \mathbf{V}^T$ where $\mathbf{\Sigma}_s$ is the $\mathbf{\Sigma}$ matrix where we keep only the leading principal submatrix of size $s \times s$ and set to zero everything else. This is a good choice since the positive diagonal elements of $\mathbf{\Sigma}$ are sorted in decreasing order of their values and therefore we expect $\mathbf{X}$ to keep entries with large magnitude. In fact, $\| \mathbf{X} \|_F^2 = \sum_{k=1}^s \sigma_k^2$, where the $\sigma_k$'s are the diagonal elements of $\mathbf{\Sigma}$, due to the fact that the rows of $\mathbf{V}^T$ have unit magnitude. Furthermore, with the same $\mathbf{X}=\mathbf{\Sigma}_s \mathbf{V}^T$ we have
$\| \mathbf{Y} - \mathbf{U X} \|_F^2 =
\| \mathbf{U} (\mathbf{\Sigma} - \mathbf{\Sigma}_s) \mathbf{V}^T \|_F^2
= \sum_{k=s+1}^n \sigma_k^2$.
We expect this error term to be relatively small since we sum over the smallest squared singular values of $\mathbf{Y}$. Therefore, with this choice of $\mathbf{U}$ and the optimal $\mathbf{X} = \mathcal{T}_s(\mathbf{U}^T\mathbf{Y})$ we have that $\| \mathbf{Y} - \mathbf{U X} \|_F^2 \leq \sum_{k=s+1}^n \sigma_k^2$, i.e., the representation error is always smaller than the error given by the best $s$-rank approximation of $\mathbf{Y}$.

In G$_m$--DLA, with the sparse representations $\mathbf{X} = \mathcal{T}_s(\mathbf{U}^T \mathbf{Y})$ we proceed to iteratively construct each G-transform. At step $k$, the problem to be solved is similar to \eqref{eq:solvemultipleGivens} but all transforms indexed above $k$ are currently the identity (not initialized) and will be computed in the following steps. 

Notice that $\mathbf{Z} = \mathbf{YX}^T$, necessary to compute all the values $C_{ij}$, is computed fully only once before the iterative process. At each iteration of the algorithms only two columns of $\mathbf{Z}$ need to be recomputed. Therefore, the update of $\mathbf{Z}$ is trivial since it involves the linear combinations of two columns according to a G-transform multiplication \eqref{eq:multiply}.
\begin{algorithm}[t]
	\caption{ \textbf{-- G$_m$--DLA. \newline
			Fast Orthonormal Transform Learning.} \newline \textbf{Input: } The dataset $\mathbf{Y} \in \mathbb{R}^{n \times N}$, the number of G-transforms $m$, the target sparsity $s$ and the number of iterations $K$. \newline \textbf{Output: } The sparsifying square orthogonal transform $\mathbf{U} = \mathbf{G}_{i_m j_m} \dots \mathbf{G}_{i_2 j_2} \mathbf{G}_{i_1 j_1}$ and sparse representations $\mathbf{X}$ such that $\| \mathbf{Y} - \mathbf{UX} \|_F^2$ is reduced.}
	\begin{algorithmic}
		\State \textbf{Initialization:}
		\begin{enumerate}
			\setlength{\itemindent}{+.25in}
			\item Perform the economy size singular value decomposition of the dataset $\mathbf{Y} = \mathbf{U} \mathbf{\Sigma} \mathbf{V}^T$.
			\item Compute sparse representations $\mathbf{X} = \mathcal{T}_s(\mathbf{U}^T \mathbf{Y})$.
			
			\item For $k = 1,\dots, m$: with $\mathbf{X}$ and all previous $k-1$ G-transforms fixed and $\mathbf{G}_{i_t j_t} = \mathbf{I},\ t = k+1,\dots,m$, construct the new $\mathbf{G}_{i_k j_k}$ where indices $(i_k, j_k)$ are given by \eqref{eq:score} and $\mathbf{\tilde{G}}_{i_k j_k} = \mathbf{UV}^T$ by the singular value decomposition ${\mathbf{J}_k}_{\{ i_k, j_k \}} = \mathbf{U \Sigma V}^T$ such that we minimize
			
		\end{enumerate}
			\begin{equation*}
			\| \mathbf{Y} -  \mathbf{G}_{i_k j_k}  \mathbf{G}_{i_{k-1} j_{k-1}} \dots \mathbf{G}_{i_1 j_1} \mathbf{X} \|_F^2 = \| \mathbf{Y} - \mathbf{G}_{i_k j_k} \mathbf{X}_k \|_F^2,
			\end{equation*}
		\begin{enumerate}
			\item[] as in \eqref{eq:solvemultipleGivens} for $\mathbf{Y}_k = \mathbf{Y}$ and $\mathbf{J}_k = \mathbf{Y} \mathbf{X}_k^T$.
		\end{enumerate}
		
		\State \textbf{Iterations} $1,\dots,K$:
		\begin{enumerate}
			\setlength{\itemindent}{+.25in}
			\item For $k = 1,\dots, m$: two-step update of $\mathbf{G}_{i_k j_k}$, with $\mathbf{X}$ and all other transforms $\mathbf{G}_{i_t j_t},\ t \neq k$ fixed, such that \eqref{eq:solvemultipleGivens} is minimized:
			\begin{enumerate}
			\setlength{\itemindent}{+.25in}
				\item Update best indices $(i_k, j_k)$ by \eqref{eq:score}.
				\item With new indices, update the transform $\mathbf{\tilde{G}}_{i_k j_k} = \mathbf{UV}^T$ by the singular value decomposition ${\mathbf{Z}_k}_{\{ i_k, j_k \}} = \mathbf{U \Sigma V}^T$, where $\mathbf{Z}_k = \mathbf{Y}_k \mathbf{X}_k^T$ as in \eqref{eq:solvemultipleGivens}.
			\end{enumerate}
			
			
			\item Compute sparse representations $\mathbf{X} = \mathcal{T}_s(\mathbf{U}^T \mathbf{Y})$, where $\mathbf{U}$ is given by \eqref{eq:factorization}.
		\end{enumerate}
	\end{algorithmic}
\end{algorithm}

\noindent\textbf{The iterations of G$_m$--DLA} update each G-transform sequentially, keeping all other constant, in order to minimize the current error term.

The algorithm is fast since the matrices involved in all computations can be updated from previous iterations. For example, at step $k+1$, notice from \eqref{eq:solvemultipleGivens} that $\mathbf{Y}_{k+1} = \mathbf{G}_{i_{k+1} j_{k+1}} \mathbf{Y}_k$ and $\mathbf{X}_{k+1} = \mathbf{G}_{i_k j_k} \mathbf{X}_k$. The same observation holds for $\mathbf{Z}_{k+1} = \mathbf{G}_{i_{k+1} j_{k+1}} \mathbf{Y}_k \mathbf{X}_k^T \mathbf{G}_{i_k j_k}^T = \mathbf{G}_{i_{k+1} j_{k+1}} \mathbf{Z}_k \mathbf{G}_{i_k j_k}^T$. Of course, $\mathbf{Y}_1 = \mathbf{G}_{i_2 j_2}^T \dots \mathbf{G}_{i_m j_m}^T \mathbf{Y}$ and $\mathbf{X}_1 = \mathbf{X}$. We always need to construct $\mathbf{Z}_1$ from scratch since $\mathbf{X}$ has been fully updated in the sparse reconstruction step.

After all transforms are computed, the dictionary $\mathbf{U}$ is never explicitly constructed. We always remember its factorization \eqref{eq:factorization} and apply it (directly or inversely) by sequentially applying the G-transforms in its composition. The total computational complexity of applying this dictionary for $\mathbf{U}^T \mathbf{Y}$ is $O(m N)$ which is $O(n \log (n) N)$ for sufficiently large $m$ (of order $n \log n$). This is to be compared with the $O(n^2 N)$ of a general orthogonal dictionary. Additionally, when consecutive G-transforms operate on different indices they can be applied in parallel, reducing the running time of G$_m$--DLA and that of manipulating the resulting dictionary.

The number of transforms $m$ could be decided during the runtime of G$_m$--DLA based on the magnitude of the largest value $C_{ij}$. Since this magnitude decides the reduction in the objective function of our problem, a threshold can be introduced to decide on the fly if a new transform is worth adding to the factorization. 

These observations are important from a computational perspective since the number of transforms is relatively high, $O(n \log n)$, and therefore their manipulation should be performed efficiently when learning the dictionary to keep the running time of G$_m$--DLA low.

Since each G-transform computed in our method maximally reduces the objective function and because the sparse reconstruction step is exact when using an orthogonal dictionary, we can guarantee that the proposed method monotonically converges to a local optimum point.

\noindent \textbf{Remark 5.} At each iteration of the proposed algorithm we update a single G-transform according to the maximum value $C_{ij}$. We have in fact the opportunity to update a maximum of $\lfloor n/2 \rfloor$ transforms simultaneously. We could for example partition the set $\{1,2,\dots,n\}$ in pairs of two and construct the corresponding G-transforms such that the sum of their $C_{ij}$ is maximized. With such a strategy fewer iterations are necessary but the problem of partitioning the indices such that the error is maximally reduced can be computationally demanding (all possible unique combinations of indices associations need to be generated). We expect G$_m$--DLA, as it is, to produce better results (lower representation error) due to the one-by-one transform update mechanism. Compared to the Householder approach \cite{DictHouseholder} we again expect G$_m$--DLA to performs better since the optimization is made over two coordinates at a time.

Even so, there are several options regarding the ordering. We can process the G-transforms in the order of their indices or in a random order for example in an effort to try to avoid local minimum points. $\hfill \blacksquare$

\noindent \textbf{Remark 6.} After indices $(i,j)$ are selected we have that $C_{ij} = 0$ and therefore this pair cannot be selected again until either index $i$ or $j$ participates in the construction of a future G-transform. This is because after constructing $\mathbf{G}_{ij}$ to minimize \eqref{eq:GivensLearning2} we have that $\mathbf{Z}_{ \{i, j\}}$ is updated to $\mathbf{Z}_{ \{i, j\}} \mathbf{\tilde{G}}_{ij}^T = \mathbf{U\Sigma V}^T \mathbf{VU}^T = \mathbf{U \Sigma U}^T$ which is symmetric and positive definite due to the solution $\mathbf{\tilde{G}}_{ij} = \mathbf{UV}^T$. $\hfill \blacksquare$

\noindent \textbf{Remark 7.} As previously discussed, the Procrustes solution $\mathbf{Q}$ is the best orthogonal minimizer of \eqref{eq:objectivewithQ}. It has been shown in \cite{DictHouseholder} that with this $\mathbf{Q}$ we have that $\mathbf{T} = \mathbf{YX}^T \mathbf{Q}^T = \mathbf{U \Sigma U}^T$ is symmetric positive semidefinite. Since $\mathbf{Q}$ is the global minimizer, there cannot be a G-transform $\mathbf{G}_{ij}$ such that $\mathbf{G}_{ij}\mathbf{Q}$ further reduces the error. This means that all symmetric $2 \times 2$ sub-matrices $\mathbf{T}_{ \{i,j\} } = \begin{bmatrix} T_{ii} & T_{ij} \\ T_{ij} & T_{jj} \end{bmatrix}$ of $\mathbf{T}$ are positive semidefinite, i.e., $T_{ii} + T_{jj} \geq 0$ and $T_{ii}T_{jj} \geq T_{ij}^2$ for all pairs $(i,j)$. This observation needs to hold for any symmetric positive definite matrix $\mathbf{T}$. Unfortunately, the converse is not true in general.

This means that even with an appropriately large $m \sim O(n^2)$, G$_m$--DLA might not always be able to match the performance of Q--DLA. This is not a major concern since in this paper we explore fast transforms and therefore $m \ll n^2$.$\hfill \blacksquare$

This concludes the presentation of the proposed G$_m$--DLA method. Based on similar principles next we provide a learning method for fast square but non-orthogonal dictionaries.

\section{A method for designing fast, general, non-orthogonal transforms: R$_m$--DLA}

In the case of orthogonal dictionaries, the fundamental building blocks like Householder reflectors and Givens rotations are readily available. This is not the case for general dictionaries. In this section we propose a building block for non-orthogonal structures in subsection A and then show how this can be used in a similar fashion to the G-transform to learn computationally efficient square non-orthogonal dictionaries by deriving the R$_m$--DLA method in subsection B.

\subsection{A building block for fast non-orthogonal transforms}

We assume no constraints on the variables $p,q,r,t$ (these are four degrees of freedom) and therefore $\mathbf{R}_{ij}$ from \eqref{eq:Rtransform} is no longer orthogonal in general. We propose to solve the following optimization problem
\begin{equation}
\underset{(i,j),\ \mathbf{\tilde{R}}_{ij}}{\text{minimize}} \ \left\| \mathbf{Y} - \mathbf{R}_{ij} \mathbf{X} \right\|_F^2.
\label{eq:GeneralLearningR}
\end{equation}
As in the G-transform case, we proceed with analyzing how indices $(i,j)$ are selected and then how to solve the optimization problem \eqref{eq:GeneralLearningR}, with the indices fixed. We define
\begin{equation}
\mathbf{Z} = \mathbf{YX}^T,\ \mathbf{W} = \mathbf{XX}^T,
\end{equation}
with entries $Z_{ij}$ and $W_{ij}$ respectively.

\noindent \textbf{Solving \eqref{eq:GeneralLearningR} for fixed $(i,j)$} leads to a least squares optimization problem as
\begin{equation}
\underset{\mathbf{\tilde{R}}_{ij}}{\text{minimize}}\ \left\| \begin{bmatrix} \mathbf{y}_i^T \\ \mathbf{y}_j^T \end{bmatrix} -
\mathbf{\tilde{R}}_{ij}
\begin{bmatrix} \mathbf{x}_i^T \\ \mathbf{x}_j^T \end{bmatrix} \right\|_F^2,
\label{eq:leastsquares}
\end{equation}
where $\mathbf{y}_i^T, \mathbf{x}_i^T$ are $i^\text{th}$ rows of $\mathbf{Y}$ and $\mathbf{X}$ respectively and whose solution is $\mathbf{\tilde{R}}_{ij} = \begin{bmatrix} Z_{ii} & Z_{ij}\\
Z_{ji} & Z_{jj} \end{bmatrix} \begin{bmatrix} W_{ii} & W_{ij} \\ W_{ji} & W_{jj}\end{bmatrix}^{-1}.$

\noindent \textbf{Choosing $(i,j)$ in \eqref{eq:GeneralLearningR}} depends on the objective function value in \eqref{eq:leastsquares} given by the least squares solution from above:
\begin{equation}
\begin{aligned}
&\left\| \begin{bmatrix} \mathbf{y}_i^T \\ \mathbf{y}_j^T \end{bmatrix} - \mathbf{\tilde{R}}_{ij}
\begin{bmatrix} \mathbf{x}_i^T \\ \mathbf{x}_j^T \end{bmatrix} \right\|_F^2 = 
\left\| \begin{bmatrix} \mathbf{y}_i^T \\ \mathbf{y}_j^T \end{bmatrix} \right\|_F^2  \\
& \  - \text{tr} \left( \begin{bmatrix} Z_{ii} & Z_{ij}\\
Z_{ji} & Z_{jj} \end{bmatrix}^T \begin{bmatrix} Z_{ii} & Z_{ij}\\
Z_{ji} & Z_{jj} \end{bmatrix} \begin{bmatrix} W_{ii} & W_{ij} \\ W_{ji} & W_{jj}\end{bmatrix}^{-1} \right).
\end{aligned}
\label{eq:objectivewithLS}
\end{equation}
This, together with the development in \eqref{eq:theRexpanded}, leads to
\begin{equation}
\begin{aligned}
\| & \mathbf{Y} - \mathbf{R}_{ij} \mathbf{X} \|_F^2 = \| \mathbf{Y} \|_F^2 + \| \mathbf{X} \|_F^2 - 2\text{tr}(\mathbf{Z}) - C_{ij}, \\
& \text{with } C_{ij} = \left\| \begin{bmatrix} \mathbf{x}_i^T \\ \mathbf{x}_j^T \end{bmatrix} \right\|_F^2 -	 2\text{tr}\left( \begin{bmatrix} Z_{ii} & Z_{ij}\\
Z_{ji} & Z_{jj} \end{bmatrix} \right) \\
& + \text{tr} \left( \begin{bmatrix} Z_{ii} & Z_{ij}\\
Z_{ji} & Z_{jj} \end{bmatrix}^T \begin{bmatrix} Z_{ii} & Z_{ij}\\
Z_{ji} & Z_{jj} \end{bmatrix} \begin{bmatrix} W_{ii} & W_{ij} \\ W_{ji} & W_{jj}\end{bmatrix}^{-1} \right).
\end{aligned}
\end{equation}
Since the matrices involved in the computation of $C_{ij}$ are $2 \times 2$ we can use the trace formula and the inversion of a $2 \times 2$ matrix formula to explicitly calculate
\begin{equation}
\begin{aligned}
C_{ij} = W_{ii} + W_{jj}& - 2(Z_{ii} + Z_{jj}) \\
+ & \frac{W_{ii}(Z_{ij}^2 + Z_{jj}^2 ) + W_{jj}(Z_{ii}^2 + Z_{ji}^2) }{W_{ii}W_{jj} - W_{ij}W_{ji}} \\
- & \frac{ (Z_{ii}Z_{ij}+Z_{ji}Z_{jj})(W_{ij}+W_{ji}) }{W_{ii}W_{jj} - W_{ij}W_{ji}} .
\end{aligned}
\label{eq:sij}
\end{equation}
Finally, to solve \eqref{eq:GeneralLearningR} we select the indices as
\begin{equation}
	(i^\star, j^\star) = \underset{j > i}{\arg \max} \ C_{ij},
	\label{eq:scoreR}
\end{equation}
and then solve a least square problem to construct $\mathbf{\tilde{R}}_{i^\star j^\star}$. The $C_{ij}$ are computed only when $W_{ii}W_{jj} - W_{ij}W_{ji} \neq 0$, otherwise $C_{ij} = -\infty$. To compute each $C_{ij}$ in \eqref{eq:sij} we need 24 operations and there are $\frac{n(n-1)}{2}$ such $C_{ij}$. The computational burden is dominated by constructing $\mathbf{Z} = \mathbf{YX}^T, \mathbf{W} = \mathbf{XX}^T$ which take $2snN$ and $snN$ operations, respectively.

\noindent \textbf{Remark 8.} A necessary condition for a dictionary $\mathbf{D} \in \mathbb{R}^{n \times n}$ to be a local minimum point for the dictionary learning problem is that all $C_{ij} = 0$ for $\mathbf{Z} = \mathbf{YX}^T\mathbf{D}^T, \mathbf{W} = \mathbf{DXX}^T \mathbf{D}^T$.$\hfill \blacksquare$

This concludes our discussion for one transform $\mathbf{R}_{ij}$. Notice that just like in the case of one G-transform, the solution given here finds the optimum $\mathbf{R}_{ij}$ to minimize \eqref{eq:GeneralLearningR}.

\subsection{A method for designing fast general transforms: R$_m$--DLA}

\begin{algorithm}[t]
	\caption{ \textbf{-- R$_m$--DLA. \newline
	Fast Non-orthogonal Transform Learning.} \newline \textbf{Input: } The dataset $\mathbf{Y} \in \mathbb{R}^{n \times N}$, the number of $\mathbf{R}_{ij}$ transforms $m$, the target sparsity $s$ and the number of iterations $K$. \newline \textbf{Output: } The sparsifying square non-orthogonal transform $\mathbf{D} = \mathbf{R}_{i_m j_m} \dots \mathbf{R}_{i_2 j_2} \mathbf{R}_{i_1 j_1} \mathbf{\Delta}$ and sparse representations $\mathbf{X}$ such that $\| \mathbf{Y} - \mathbf{DX} \|_F^2$ is reduced.}
	\begin{algorithmic}
		\State \textbf{Initialization:}
		\begin{enumerate}
			\setlength{\itemindent}{+.25in}
			\item Perform the economy size singular value decomposition of the dataset $\mathbf{Y} = \mathbf{U} \mathbf{\Sigma} \mathbf{V}^T$.
			\item Compute sparse representations $\mathbf{X} = \mathcal{T}_s(\mathbf{U}^T \mathbf{Y})$.

		\end{enumerate}
		
		\State \textbf{Iterations} $1,\dots,K$:
		\begin{enumerate}
			\setlength{\itemindent}{+.25in}
			\item For $k = 1,\dots, m$: with $\mathbf{X}$ and all previous $k-1$ R-transforms fixed and $\mathbf{R}_{i_t j_t} = \mathbf{I},\ t = k+1,\dots,m$, construct the new $\mathbf{R}_{i_k j_k}$ where indices $(i_k, j_k)$ are given by \eqref{eq:scoreR} and $\mathbf{\tilde{R}}_{i_k j_k}$ is given by the least squares solution that minimizes \eqref{eq:solvemultipleR}.
			
			\item Compute $\mathbf{\Delta}$ in \eqref{eq:factorizationofD} such that $\| \mathbf{d}_j \|_2 = 1$.
			
			\item Compute sparse representations $\mathbf{X} \! \! = \! \! \text{OMP}(\mathbf{D}, \mathbf{Y}, s)$ where $\mathbf{D}$ is given in \eqref{eq:factorizationofD}.
		\end{enumerate}
		
				\State \textbf{Iterations} $1,\dots,K$:
				\begin{enumerate}
					\setlength{\itemindent}{+.25in}
					\item For $k = 1,\dots, m$: with $\mathbf{X}$, indices $(i_k, j_k)$ and all transforms except the $k^\text{th}$ fixed, update only the non-zero part of $\mathbf{R}_{i_k j_k}$, denoted $\mathbf{\tilde{R}}_{i_k j_k}$, such that \eqref{eq:foriterationsR} is minimized.
					
					\item Compute $\mathbf{\Delta}$ in \eqref{eq:factorizationofD} such that $\| \mathbf{d}_j \|_2 = 1$.
					
					\item Compute sparse representations $\mathbf{X} \! \! = \! \! \text{OMP}(\mathbf{D}, \mathbf{Y}, s)$ where $\mathbf{D}$ is given in \eqref{eq:factorizationofD}.
				\end{enumerate}
	\end{algorithmic}
\end{algorithm}

Similarly to G$_m$--DLA, we now propose to construct a general dictionary $\mathbf{D} \in \mathbb{R}^{n \times n}$ with the following structure:
\begin{equation}
\mathbf{D} = \mathbf{R}_{i_m j_m} \dots \mathbf{R}_{i_2 j_2} \mathbf{R}_{i_1 j_1} \mathbf{\Delta}.
\label{eq:factorizationofD}
\end{equation}
The value of $m$ is a choice of the user. For example, if we choose $m$ to be $O(n \log n)$ the dictionary $\mathbf{D}$ can be applied in $O(n \log n)$ computational complexity -- similar to the classical fast transforms. The goal of this section is to propose a learning method that constructs such a general dictionary. As the transformations $\mathbf{R}_{ij}$ are general, the diagonal matrix $\mathbf{\Delta} \in \mathbf{R}^{n \times n}$ is there to ensure that all columns $\mathbf{d}_j$ of $\mathbf{D}$ are normalized $\| \mathbf{d}_j \|_2 = 1$ (as in the formulation \eqref{eq:dictionaryGeneral}). This normalization does not affect the performance of the method since $\mathbf{D}\mathbf{X}$ is equivalent to $(\mathbf{D} \mathbf{\Delta}) (\mathbf{\Delta}^{-1} \mathbf{X})$.

We fix the representations $\mathbf{X}$ and all transforms in \eqref{eq:factorizationofD} except for the $k^\text{th}$ transform $\mathbf{R}_{i_k j_k}$. Moreover, all transforms $\mathbf{R}_{i_{k+1} j_{k+1}}, \dots, \mathbf{R}_{i_m j_m}$ are set to $\mathbf{I}$. Because the transforms $\mathbf{R}_{ij}$ are not orthogonal we cannot access directly any transform $\mathbf{R}_{i_k j_k}$ in \eqref{eq:factorizationofD}, but only the left most one $\mathbf{R}_{i_m j_m}$. In this case, to optimize the dictionary $\mathbf{D}$ only for this $\mathbf{R}_{i_k j_k}$ transform we reach the objective
\begin{equation}
\| \mathbf{Y} - \mathbf{R}_{i_k j_k} \dots \mathbf{R}_{i_2 j_2} \mathbf{R}_{i_1 j_1} \mathbf{X} \|_F^2 =  \| \mathbf{Y}  - \mathbf{R}_{i_k j_k} \mathbf{X}_k \|_F^2.
\label{eq:solvemultipleR}
\end{equation}
Therefore, our goal is to solve
\begin{equation}
	\underset{\mathbf{R}_{i_k j_k}}{\text{minimize}}\ \| \mathbf{Y} - \mathbf{R}_{i_k j_k} \mathbf{X}_k \|_F^2.
\end{equation}

Notice that we have reduced the problem to the formulation in \eqref{eq:GeneralLearningR} whose full solution is outlined in the previous section. We can apply this procedure for all G-transforms in the product of $\mathbf{D}$ and therefore a full update procedure presents itself: we will sequentially update each transform in \eqref{eq:factorizationofD}, from the right to the left, and then the sparse representations until convergence or for a total number of iterations $K$.

Once these iterations terminate we can refine the result. As previously mentioned, we cannot arbitrarily update a transform $\mathbf{\tilde{R}}_{i_k j_k}$ because this transform is not orthogonal. But we can update its non-zero part $\mathbf{\tilde{R}}_{i_k j_k}$. Consider the development:
\begin{equation}
	\begin{aligned}
	\| & \mathbf{Y} \! \! - \! \! \mathbf{R}_{i_m j_m} \dots \mathbf{R}_{i_{k+1} j_{k+1}} \mathbf{R}_{i_k j_k} \mathbf{R}_{i_{k-1} j_{k-1}} \dots \mathbf{R}_{i_1 j_1} \mathbf{X} \|_F^2 \\
	= & \| \mathbf{Y} - \mathbf{B}_k \mathbf{R}_{i_k j_k} \mathbf{X}_k \|_F^2 \\
	= & \| \text{vec}(\mathbf{Y}) - (\mathbf{X}_k^T \otimes \mathbf{B}_k) \text{vec}(\mathbf{R}_{i_k j_k}) \|_F^2 \\
	= & \left\| \text{vec}(\mathbf{Y}) - \! \! \! \! \! \sum_{t \in \{1,\dots,n\} \setminus \{ i_k, j_k \}} \! \! \! \! \! \! \! \! ( (\mathbf{X}_k^T)_t \otimes (\mathbf{B}_k)_t) -  \mathbf{Cx} \right\|_F^2 \\
	= & \| \mathbf{w} - \mathbf{Cx} \|_F^2,
	\end{aligned}
	\label{eq:foriterationsR}
\end{equation}
where $\mathbf{x} = \text{vec}(\mathbf{\tilde{R}}_{i_k j_k}) \in \mathbb{R}^4$ and $\mathbf{C} = [ (\mathbf{X}_k^T)_{i_k} \otimes (\mathbf{B}_k)_{i_k}  \ \ (\mathbf{X}_k^T)_{i_k} \otimes (\mathbf{B}_k)_{j_k} \ \ 
   (\mathbf{X}_k^T)_{j_k} \otimes (\mathbf{B}_k)_{i_k} \ \  (\mathbf{X}_k^T)_{j_k} \otimes (\mathbf{B}_k)_{j_k} ] \in \mathbb{R}^{nN \times 4}$.
We have denoted by $(\mathbf{B}_k)_{i_k}$ the $i_k^\text{th}$ column of $\mathbf{B}_k$ and $\otimes$ is the Kronecker product. To develop \eqref{eq:foriterationsR} we have used the fact that the Frobenius norm is an elementwise operator, the structure of $\mathbf{R}_{i_k j_k}$ and the fact that
\begin{equation}
 \text{vec}(\mathbf{B}_k \mathbf{R}_{i_k j_k} \mathbf{X}_k) = (\mathbf{X}_k^T \otimes \mathbf{B}_k) \text{vec}(\mathbf{R}_{i_k j_k}).
\end{equation}
The $\mathbf{x}$ that minimizes \eqref{eq:foriterationsR} is given by the least squares solution $\mathbf{x} = (\mathbf{C}^T \mathbf{C})^{-1} \mathbf{C}^T \mathbf{w}$. Therefore, once the product of the $m$ transforms is constructed we can update the non-zero part of any transform to further reduce the objective function. What we cannot do is update the indices $(i_k, j_k)$ on which the calculation takes place, these stay the same.

Therefore, we propose a learning procedure that has two sets of iterations: the first constructs the transforms $\mathbf{R}_{i_k j_k}$ in a rigid manner, ordered from right to left most, and the second only updates the non-zero parts $\mathbf{\tilde{R}}_{i_k j_k}$ of all the transforms without changing the coordinates $(i_k, j_k)$. The full procedure we propose, called R$_m$--DLA, is detailed in Algorithm 2. This algorithm has two main parts which we will now describe.

\noindent\textbf{The initialization of R$_m$--DLA} has the goal to construct the sparse representation matrix $\mathbf{X} \in \mathbb{R}^{n \times N}$. We have several options in this step. We can construct $\mathbf{X}$ in the same way as for G$_m$--DLA from the singular value decomposition of the dataset or by running another dictionary learning algorithm (like the K--SVD \cite{KSVD} for example) and use the $\mathbf{X}$ it constructs.

\noindent\textbf{The iterations of R$_m$--DLA} are divided into two sets. The goal of the first set of iterations is to decide upon all the indices $(i_k, j_k)$ while the second set of iterations optimizes over the non-zero components of all the transforms in the factorization with the fixed indices previously decided.

The proposed R$_m$--DLA is can be itself efficiently implemented. When iteratively solving problems as \eqref{eq:solvemultipleR} we have that $\mathbf{X}_{k+1} = \mathbf{R}_{i_k j_k} \mathbf{X}_k$ with $\mathbf{X}_1 = \mathbf{X}$ while when iteratively solving problems as \eqref{eq:foriterationsR} we have that $\mathbf{X}_{k+1} = \mathbf{R}_{i_k j_k} \mathbf{X}_k$ and $\mathbf{B}_{k+1} = \mathbf{B}_k \mathbf{R}_{i_k j_k}^{-1}$ with $\mathbf{X}_1 = \mathbf{X}$ and $\mathbf{B}_1 = \mathbf{R}_{i_m j_m}\dots\mathbf{R}_{i_2 j_2}$. The explicit inverse $\mathbf{R}_{i_k j_k}^{-1}$ is not computed, instead the equivalent linear system for 2 variables can be efficiently solved.

The updates of all the transforms $\mathbf{R}_{i_k j_k}$ monotonically decrease our objective function since we solve exactly the optimization problems in these variables. Unfortunately, normalizing to unit $\ell_2$ norm the columns of the transform and constructing the sparse approximations via OMP, which is not an exact optimization step, may cause increases in the objective function. For these reasons, monotonic convergence of R$_m$--DLA to a local minimum point cannot be guaranteed. For this reason, at all times we keep track of the best solution pair $(\mathbf{D}, \mathbf{X})$ and return it at the end of each iterative process.


This concludes our discussion of R$_m$--DLA. We now move to discuss the computational complexity of the transforms created by the proposed methods and to show experimentally their representation capabilities.

\section{The computational complexity of using learned transforms}

In this section we look at the computational complexity of using the learned dictionaries to create the sparse representations on a dataset $\mathbf{Y}$ of size $n \times N$. We are in a computational regime where we assume dimensions obey
\begin{equation}
	s \ll n \ll N.
\end{equation}

The computational complexity of using a general non-orthogonal dictionary $\mathbf{A}$ of size $n \times n$ in sparse recovery problems with Batch--OMP \cite{AKSVD} is
\begin{equation}
N_\mathbf{A} \approx (2n^2 + s^2n  + 3sn + s^3)N + n^3.
\label{eq:NOMPCholesky}
\end{equation}
The cost of $n^3$ is associated with the construction of the Gram matrix of the dictionary and it does not depend on the number of samples $N$ in the data. The total number of operations is dominated by constructing the projections in the dictionary column space which takes $2n^2$ operations per sample. The other operations dependent on the sparsity $s$ and express the cost of iteratively finding the support of the sparse approximation.

The computational complexity of using an orthogonal dictionary $\mathbf{Q}$ designed via Q--DLA is
\begin{equation}
	N_\mathbf{Q} \approx (2n^2 + ns)N.
	\label{eq:NOMPCholeskyQ}
\end{equation}
As in the general case, the cost is dominated by constructing the projections $\mathbf{Q}^T \mathbf{Y}$ which takes $2n^2$ operations for each of the $N$ columns in $\mathbf{Y}$. The cost of $ns$ expresses the approximate work done to identify the largest $s$ entries in magnitude in the representation of each data sample. This can be performed in an efficient manner by keeping the $s$ largest components in magnitude while the projections are computed for each data sample. Compared with \eqref{eq:NOMPCholesky}, the iterative steps of constructing the support of the OMP solution for each sample and the construction of the Gram matrix (which is the identity matrix in this case) is no longer needed.

The same operation with a dictionary $\mathbf{U}$ as \eqref{eq:factorization} computed via G$_{m_1}$--DLA takes
\begin{equation}
N_\mathbf{U} \approx (6m_1 + ns)N.
\label{eq:NOMPCholeskyForF}
\end{equation}
The result is similar to \eqref{eq:NOMPCholeskyQ} but now the cost of constructing the projections $\mathbf{U}^T \mathbf{Y}$ takes now only $6m_1$ operations per data sample. Here is where the G-transform factorization is used explicitly to reduce the computational complexity.

Finally, with a dictionary $\mathbf{D}$ as \eqref{eq:factorizationofD} computed via R$_{m_2}$--DLA the sparse approximation step via Batch--OMP \cite{AKSVD} takes
\begin{equation}
N_\mathbf{D} \approx (6m_2 + n + s^2n + 3sn + s^3)N + 6m_2n.
\label{eq:NOMPCholeskyForR}
\end{equation}
In this case, the cost of building the projections $\mathbf{D}^T \mathbf{Y}$ takes $6m_2$ operations to apply each $\mathbf{R}_{ij}$ transform and then $n$ operations to apply the scaling of the diagonal $\mathbf{\Delta}$. Simplifications occur also for the construction of the symmetric Gram matrix $\mathbf{D}^T \mathbf{D}$ which now takes $6m_2n$ operations, instead of the regular $n^3$ operations. This later simplification might not be significant since it is not dependent on the size of the dataset $N$.

A dictionary $\mathbf{U}$ designed via G$_{m_1}$--DLA has approximately the same computationally complexity as a general orthogonal dictionary $\mathbf{Q}$ designed via Q--DLA when
\begin{equation}
	m_1 = \left\lfloor \frac{n^2}{3} \right\rfloor.
	\label{eq:limitU}
\end{equation}
Because any orthogonal matrix can be factorized as a product of $\frac{n(n-1)}{2}$ G-transforms and because of the upper limit imposed in \eqref{eq:limitU} it is clear that we cannot express any orthogonal dictionary as an efficient transform for sparse representations. In some cases, the full orthogonal dictionary $\mathbf{Q}$ might be more efficient than its factorization with G-transforms. In general, the representation error achieved by general orthogonal dictionaries designed via Q--DLA is a performance limit for G-transform based dictionaries.

A similar comparison can be made between the computational complexity of a general dictionary $\mathbf{A}$ and that of a dictionary $\mathbf{D}$ composed of $m_2$ transformations $\mathbf{R}_{ij}$. Their complexities approximately match when
\begin{equation}
	m_2 = \left\lfloor \frac{(2n^2 +s^2n + 3sn + s^3)N + n^3}{6(N+n)} \right\rfloor \! \! \! \overset{N \to \infty}{\approx} \! \! \! \left\lfloor \frac{n^2}{3} \right\rfloor.
\end{equation}

This shows that for both G$_m$--DLA and R$_m$--DLA the computationally efficient regimes are when $m \sim O(n)$ or in general $m \ll n^2$.

A last comment regards the comparison between dictionaries created with G$_{m_1}$--DLA and R$_{m_2}$--DLA. When $m_1 = m_2$ we expect R$_{m_2}$--DLA to perform better but at a higher computational cost. Assuming large datasets $N \to \infty$ and low sparsity $s \ll n$, computational complexities approximately match when
\begin{equation}
	m_1 \approx \left\lfloor m_2 + \frac{(s^2 + 3s + 1)n}{6} \right\rfloor.
	\label{eq:m1m2}
\end{equation}

Due to the use of the OMP procedure for non-orthogonal dictionaries to create the sparse approximations, dictionaries designed via R$_m$--DLA are much more computationally complex than the orthogonal dictionaries designed via G$_m$--DLA. Otherwise, as depicted in \eqref{eq:m1m2}, for the same representation performance the orthogonal dictionaries may contain many more G-transforms in their factorization than $\mathbf{R}_{ij}$ transforms contained in the factorization of a non-orthogonal dictionary. As a consequence, it may be that orthogonal dictionaries are always more computationally efficient than general dictionaries for approximately equal representation capabilities. A definite advantage of R$_m$--DLA is that it has the potential to create dictionaries that go below representation errors given by orthogonal dictionaries designed via Q--DLA, the performance limit of G$_m$--DLA.

Using these approximate complexities, we discuss in the results section the representation performance versus the computational complexity trade-off that the dictionaries constructed via the proposed methods display.

\section{Experimental results}

In this section we provide experimental results that show how transforms designed via the proposed methods G$_m$--DLA and R$_m$--DLA behave on image data.


\begin{figure}[t]
	\centering
	\includegraphics[trim = 18 0 30 15, clip, width=0.3\textwidth]{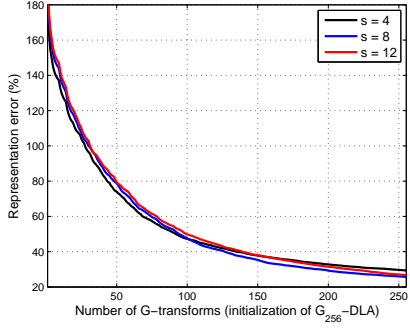}
	\caption{For the proposed G$_{256}$--DLA we show the relative representation error \eqref{eq:reperror} in the initialization steps for the dataset $\mathbf{Y}$ created from the patches of the images couple, peppers and boat with sparsity $s \in \{4, 8, 12\}$. Notice that in general the representation error can surpass 100\%, for example, for orthogonal dictionaries, the maximum value $\epsilon = 4$ is achieved when $\mathbf{X} = -\mathbf{Y}$ and $\mathbf{D} = \mathbf{I}$ in \eqref{eq:reperror}.}
	\label{fig:InitializationEvolution}
\end{figure}

\begin{figure}[t]
	\centering
	\includegraphics[trim = 18 7 30 20, clip, width=0.3\textwidth]{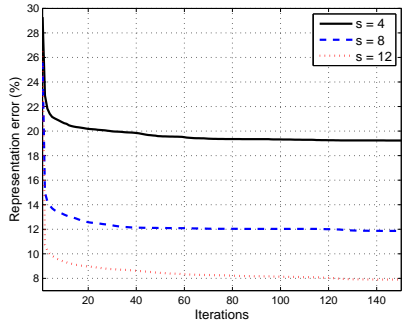}
	\caption{For the same experimental setup as in Figure \ref{fig:InitializationEvolution}, we show the representation error for the $K=150$ regular iterations of G$_{256}$--DLA.}
	\label{fig:IterationEvolution}
\end{figure}

\begin{figure}[t]
	\centering
	\includegraphics[trim = 18 7 30 20, clip, width=0.3\textwidth]{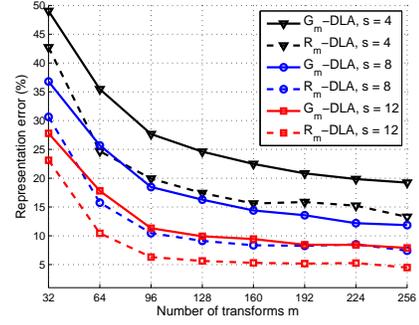}
	\caption{Performance of G$_m$--DLA and R$_m$--DLA in terms of the relative representation error \eqref{eq:reperror} for different sparsity levels $s \in \{4, 8, 12\}$.}
	\label{fig:IterationWithSandM}
\end{figure}

\begin{figure}[t]
	\centering
	\includegraphics[trim = 18 7 30 20, clip, width=0.3\textwidth]{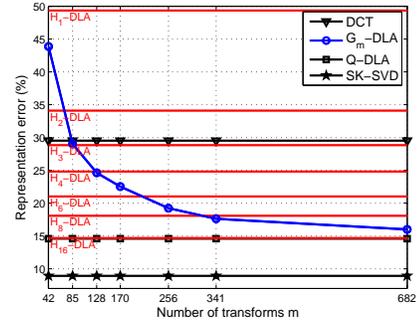}
	\caption{Comparisons, in terms of relative representation errors \eqref{eq:reperror}, of G$_m$--DLA against the DCT, Q--DLA \cite{OrthoDictionary}, SK--SVD \cite{SKSVD} and Householder based orthogonal dictionaries \cite{DictHouseholder} denoted here H$_p$--DLA where $p$ is the number of reflectors in the factorization of the dictionary. The number of transforms $m$ is chosen so that computational complexity comparisons against H$_p$--DLA is possible. Computational complexity approximately match between: H$_1$--DLA and G$_{42}$--DLA, H$_2$--DLA and G$_{85}$--DLA, H$_3$--DLA and G$_{128}$--DLA, H$_4$--DLA and G$_{170}$--DLA, H$_6$--DLA and G$_{256}$--DLA, H$_8$--DLA and G$_{341}$--DLA, H$_{16}$--DLA and G$_{682}$--DLA. The sparsity level is set to $s = 4$ for all methods. We use the SK--SVD to build a square, non-orthogonal, dictionary.}
	\label{fig:Comparisons}
\end{figure}

\begin{figure}[t]
	\centering
	\includegraphics[trim = 18 7 30 20, clip, width=0.3\textwidth]{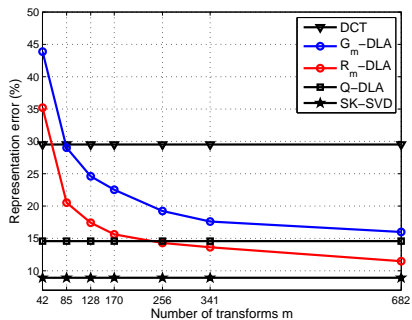}
	\caption{For the same experimental setup as in Figure \ref{fig:Comparisons} we compare G$_m$--DLA against R$_m$--DLA.}
	\label{fig:Comparisons2}
\end{figure}

\begin{figure}[t]
	\centering
	\includegraphics[trim = 18 7 30 20, clip, width=0.3\textwidth]{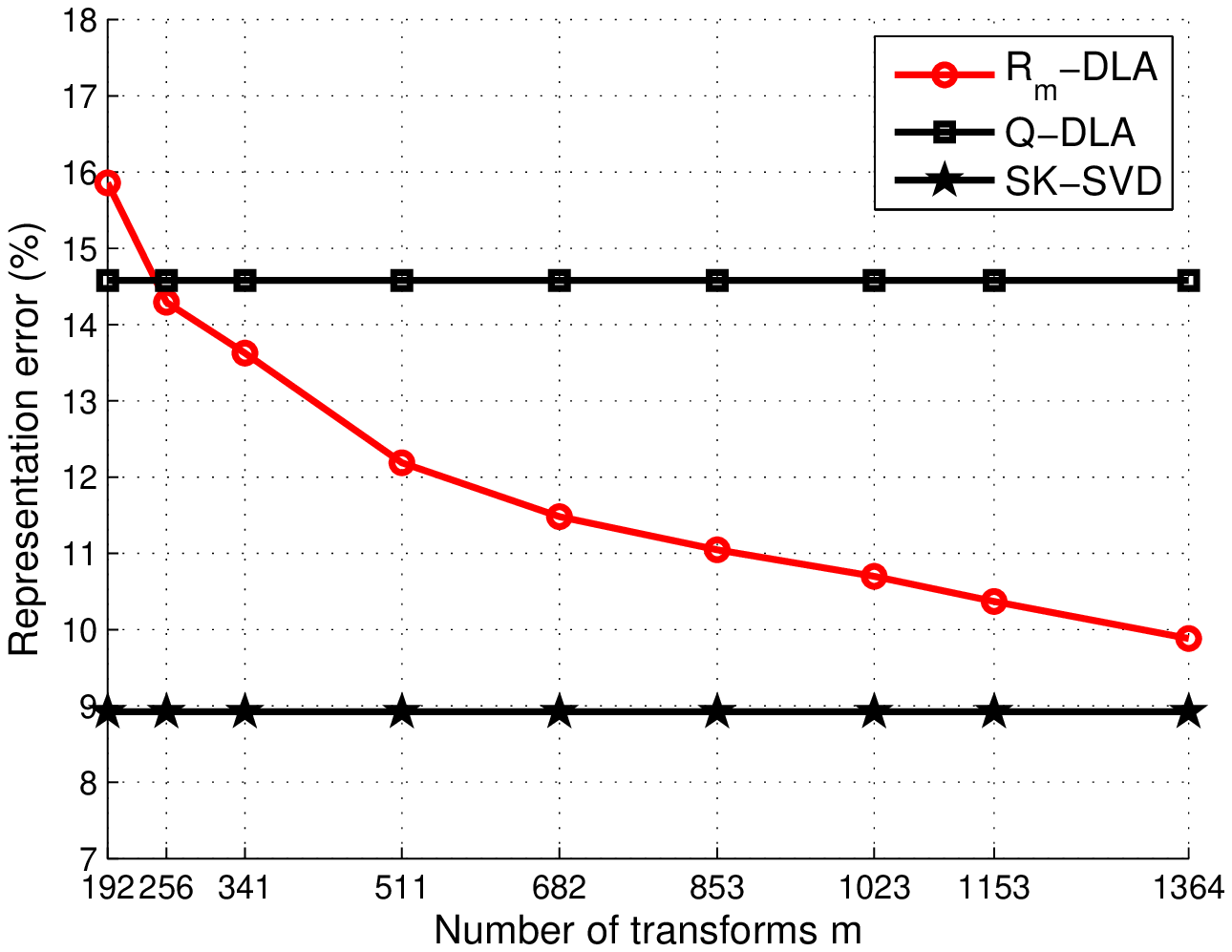}
	\caption{The evolution of R$_m$--DLA for $m$ large enough to outperform any orthogonal dictionary.}
	\label{fig:Comparisons3}
\end{figure}

The input data that we consider are taken from popular test images from the image processing literature (pirate, peppers, boat etc.). The test dataset $\mathbf{Y} \in \mathbb{R}^{n \times N}$ consists of $8 \times 8$ non-overlapping patches with their means removed and normalized $\mathbf{Y} \leftarrow \mathbf{Y}/255$. We choose to compare the proposed methods on image data since in this setting fast transforms that perform very well, like the Discrete Cosine Transform (DCT) \cite{FCT} for example, are available. Our goal is to provide dictionaries based on factorizations like \eqref{eq:factorization} and \eqref{eq:factorizationofD} that perform well in terms of representation error with a small number $m$ of basic transforms in their composition. All algorithms run for $K = 150$ iterations and there are $N = 12288$ image patches in the dataset $\mathbf{Y}$ each of size $n = 64$.

To measure the quality of a dictionary $\mathbf{D}$ we choose to evaluate the relative representation error
\begin{equation}
	\epsilon = \| \mathbf{Y} - \mathbf{DX} \|_F^2 \| \mathbf{Y} \|_F^{-2}\ (\%).
	\label{eq:reperror}
\end{equation}

Figures \ref{fig:InitializationEvolution} and \ref{fig:IterationEvolution} show the evolution of G$_{256}$--DLA for $K = 150$ iterations (including the initialization procedure, i.e., the first 256 steps of the algorithm). The figures show how effective the initialization is in reducing the representation error for any sparsity level. Notice that the initialization procedure provides similar results regardless of the sparsity level $s$. The $K= 150$ iterations of G$_{256}$--DLA further lower the representation error providing better results with larger sparsity level. As previously discussed, each step of the algorithm monotonically decreases the objective function of the dictionary learning problem until convergence.

Figure \ref{fig:IterationWithSandM} shows how G$_m$--DLA evolves with the number of transforms $m$ and the sparsity $s$. As expected, increasing the number of transforms $\mathbf{G}_{ij}$ and $\mathbf{R}_{ij}$ in the factorization always lowers the representation error but with diminishing returns as $m$ increases. This figure helps choose the number of transforms $m$ while balancing between the computational complexity and representation performance. Large decreases in the representation error are seen up to $m=96$ or $m=128$ while thereafter increasing $m$ brings smaller benefits. Also, with higher sparsity levels the number of transforms $m$ becomes less relevant. We notice that with $s=12$ the representation performance hits a plateau after $m \geq 128$ transforms.

An interesting point of comparison is between the dictionaries constructed via G$_m$--DLA and H$_p$--DLA \cite{DictHouseholder}. Figure \ref{fig:Comparisons} provides a detailed comparison between the two. A matrix-vector multiplication takes $4n$ operations for a reflector and only $6$ operations for a G-transform. If we compare the computational complexities of the dictionaries constructed by the two methods we find approximate equality between H$_p$--DLA and G$_{\left\lfloor \frac{2}{3}np \right\rfloor}$--DLA. Notice from this figure that for a low $m$ the G-transform approach provides better results than the Householder approach while also enjoying lower computational complexity. As the complexity of the dictionaries increases (larger number of G-transforms or reflectors) the gap between the two approaches decreases. The most complex dictionaries are designed via H$_{16}$--DLA and G$_{682}$--DLA and they closely match the performance of the general orthogonal dictionary learning approach Q--DLA while still keeping a computational advantage. In this case, the Householder approach keeps a slight edge in representation performance. Since the proposed approach updates the G-transforms sequentially the probability of getting stuck in local minimum points is more likely with large $m$. The difficulties that G$_m$--DLA encounters for large $m$ are also discussed in Remark 6.

It is also interesting to see that the representation performance of the DCT is matched by H$_{3}$--DLA and G$_{85}$--DLA. The computational complexity of H$_3$--DLA approximately matches that of the DCT \cite{FCT} (based on the FFT) while G$_{85}$--DLA is actually computationally simpler than the DCT. In fact, any dictionary constructed by G$_m$--DLA for $85 \leq m \leq 128$ is faster and provides better representations than the DCT.

Figure \ref{fig:Comparisons2} compares the G$_m$--DLA against the R$_m$--DLA for the same number of transforms $m$ in their factorizations. R$_m$--DLA always outperforms G$_m$--DLA since the $\mathbf{G}_{ij}$ is a constrained version of $\mathbf{R}_{ij}$. Unfortunately, the non-orthogonal transforms also have much higher computational complexity than their orthogonal counterparts in the sparse approximation step. For example, the computational complexity is approximately equivalent between dictionaries designed via R$_{42}$--DLA and G$_{351}$--DLA. The main benefit of non-orthogonal transforms is that ultimately, for large enough $m$, their performance surpasses that of general (computationally inefficient) orthogonal transforms designed via Q--DLA. In our case this happens for $m \geq 256$. The performance of the DCT is approximately matched by R$_{50}$--DLA. Surprisingly, less than $n$ factors in the product of the transform suffice to match the performance of the classical DCT transform for sparse recovery. This highlights the way dictionaries designed via R$_m$--DLA balance the computationally efficiency and representation performance trade-off, i.e., one R-transform gives 4 degrees of freedom for the cost of 6 operations.

Figure \ref{fig:Comparisons3} shows the performance of R$_m$--DLA in a regime close to the results of the SK--SVD dictionary learning method \cite{SKSVD}. We use the SK--SVD to construct a square dictionary, i.e., a dictionary with $n$ atoms. The complexity of the dictionary designed via R$_{1364}$--DLA matches that of the dictionary designed via SK--SVD while there is a small performance gap between the two. We notice experimentally that the iterative procedure of R$_m$--DLA improves performance always when increasing $m$ but the probability of getting stuck in local minimum points increases. Therefore, just as G$_m$--DLA has some trouble matching the performance of Q--DLA, R$_m$--DLA has trouble exactly matching the performance of SK--SVD.

\begin{figure}[t]
	\centering
	\includegraphics[trim = 18 7 25 20, clip, width=0.3\textwidth]{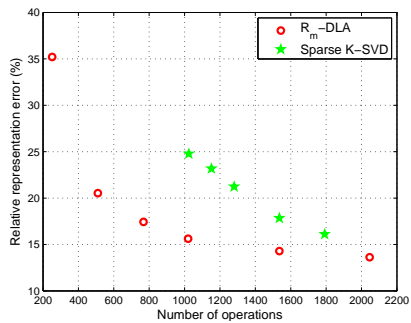}
	\caption{Pareto curves for R$_m$--DLA and the Sparse K--SVD approach \cite{SparseKSVD}. We consider the representation error in \eqref{eq:reperror} and the number of operations necessary to perform $\mathbf{D}^T \mathbf{y}$ given a target vector $\mathbf{y} \in \mathbb{R}^n$. We train five square dictionaries $\mathbf{D} = \mathbf{\Phi S}$ with the Sparse K--SVD approach, each with a different sparsity parameter $p \in \{2,3,4,6,8\}$ in the matrix $\mathbf{S}$. For a transform created with R$_m$--DLA matrix-vector multiplication takes $6m$ operations. The experimental setup for training the transforms is the same as in Figure \ref{fig:Comparisons}.}
	\label{fig:VSSparseKSVD}
\end{figure}

In the last experimental setup we compare our R$_m$--DLA with the previously proposed Sparse K--SVD approach \cite{SparseKSVD}. We use the Sparse K--SVD to build a square dictionary $\mathbf{D} = \mathbf{\Phi}\mathbf{S} \in \mathbb{R}^{n \times n}$ where $\mathbf{\Phi}$ is a well-known classic transform (in our case the DCT) and $\mathbf{S} \in \mathbb{R}^{n \times n}$ is matrix with only $p$ non-zero entries per column. In this fashion, matrix-vector multiplication like $\mathbf{D}^T \mathbf{y} = \mathbf{S}^T\mathbf{\Phi}^T \mathbf{y}$ takes $2pn + C$ operations, where $C$ is the cost of applying the DCT (in our case, this is the same as using a transform designed via G$_{128}$--DLA or R$_{128}$--DLA). R$_m$--DLA performs consistently better than the Sparse K--SVD for very fast transforms while the performance gap closes for very low representation errors. The Sparse K--SVD suffers from the fact that the fast transform $\mathbf{\Phi}$ is fixed and therefore the optimization takes place over only $pn$ degreed of freedom. We restrict ourselves to square transforms and avoid the comparison with overcomplete dictionaries designed via the K--SVD or the Sparse K--SVD. Experimental insights into how the representation performance scales with the number of atoms in the dictionary are given in \cite{SKSVD,RusuDictionaries2012}.

When designing a very fast orthogonal transform (whose complexity let us say is order $n$ or $n \log n$) then G$_m$--DLA provides very good results while achieving the lowest computational complexity. For improved performance, more complex orthogonal transforms perform better when designed via H$_p$--DLA. If representation capabilities is the only performance metric then the non-orthogonal transforms designed by R$_m$--DLA are the weapon of choice. For large $m$ both G$_m$--DLA and R$_m$--DLA can suffer from long running times. For example, G$_{128}$--DLA takes several minutes to terminate while R$_{128}$--DLA's running time is close to ten minutes on a modern Intel i7 computer system. We note that the algorithms are implemented in Matlab$^\text{\textregistered}$. A careful implementation in a lower level compiled programming language will drive these running times much lower and reduce the memory footprint.

\section{Conclusions}

In this paper we present practical procedures to learn square orthogonal and non-orthogonal dictionaries already factored into a fixed number of computationally efficient blocks. We show how effective the dictionaries constructed via the proposed methods are on image data where we compare against the fast cosine transform on one side and general non-orthogonal and orthogonal dictionaries on the other. We also show comparisons with a recently proposed method that constructs Householder based orthogonal dictionaries. We show empirically that the proposed methods construct transforms that provide an improved trade-off between computational complexity and representation performance among the methods we consider. We are able to construct transforms that exhibit lower computational efficiency and lower representation error than the fast cosine transform for image data. We expect the current work to extend the use of learned transforms in time critical scenarios and to devices where, due to power limitations, only low complexity algorithms can be deployed.

\bibliographystyle{IEEEtran}
\bibliography{refs}

\end{document}